\documentclass[sigconf]{acmart}
\AtBeginDocument{%
	\providecommand\BibTeX{{%
			\normalfont B\kern-0.5em{\scshape i\kern-0.25em b}\kern-0.8em\TeX}}}
\usepackage{balance}
\setcopyright{acmcopyright}
\copyrightyear{2022}
\acmYear{2022}
\acmDOI{10.1145/3503161.3548114}

\acmConference[MM '22] {Proceedings of the 30th ACM International Conference on Multimedia }{October 10--14, 2022}{Lisboa, Portugal.}
\acmBooktitle{Proceedings of the 30th ACM International Conference on Multimedia (MM '22), October 10--14, 2022, Lisboa, Portugal}  
\acmPrice{15.00}
\acmISBN{978-1-4503-9203-7/22/10}

\begin{document}
	%\fancyhead{}
	
	\title{Exploring High-quality Target Domain Information for Unsupervised Domain Adaptive Semantic Segmentation}
	
	\author{Junjie Li}
	\email{hnljj@mail.ustc.edu.cn}
	\orcid{0000-0002-2906-8514}
	%\email{webmaster@marysville-ohio.com}
	\affiliation{%
		\institution{University of Science and Technology of China}
		%\streetaddress{P.O. Box 1212}
		\city{Hefei}
		\country{China}
	}
	
	\author{Zilei Wang}
	\authornote{Corresponding Author}
	\email{zlwang@ustc.edu.cn}
	\orcid{0000-0003-1822-3731}
	\affiliation{%
		\institution{University of Science and Technology of China}
		%\streetaddress{1 Th{\o}rv{\"a}ld Circle}
		\city{Hefei}
		\country{China}
	}

	\author{Yuan Gao}
	\email{gyy@mail.ustc.edu.cn}
	\orcid{0000-0003-3899-0698}
	\affiliation{%
		\institution{University of Science and Technology of China}
		\city{Hefei}
		\country{China}
	}
	
	\author{Xiaoming Hu}
	\email{cjdc@mail.ustc.edu.cn}
	\orcid{0000-0002-4853-7814}
	\affiliation{%
		\institution{University of Science and Technology of China}
		%\streetaddress{Rono-Hills}
		\city{Hefei}
		%\state{Arunachal Pradesh}
		\country{China}
	}

	%\renewcommand{\shortauthors}{Junjie Li, et al.}
\renewcommand{\shortauthors}{Junjie Li, Zilei Wang, Yuan Gao, \& Xiaoming Hu}	
	\begin{abstract}
		In unsupervised domain adaptive (UDA) semantic segmentation, the distillation based methods are currently dominant in performance. However, the distillation technique requires complicate multi-stage process and many training tricks. 
		In this paper, we propose a simple yet effective method that can achieve competitive performance to the advanced distillation methods. Our core idea is to fully explore the target-domain information from the views of boundaries and features. 
		First, we propose a novel mix-up strategy to generate high-quality target-domain boundaries with ground-truth labels. Different from the source-domain boundaries in previous works, we select the high-confidence target-domain areas and then paste them to the source-domain images. Such a strategy can generate the object boundaries in target domain (edge of target-domain object areas) with the correct labels. Consequently, the boundary information of target domain can be effectively captured by learning on the mixed-up samples.
		Second, we design a multi-level contrastive loss to improve the representation of target-domain data, including pixel-level and prototype-level contrastive learning. 
		By combining two proposed methods, more discriminative features can be extracted and hard object boundaries can be better addressed for the target domain.
		The experimental results on two commonly adopted benchmarks (\textit{i.e.}, GTA5 $\rightarrow$ Cityscapes and SYNTHIA $\rightarrow$ Cityscapes) show that our method achieves competitive performance to complicated distillation methods. Notably, for the SYNTHIA$\rightarrow$ Cityscapes scenario, our method
		achieves the state-of-the-art performance with $57.8\%$ mIoU and $64.6\%$ mIoU on 16 classes and 13 classes. Code is available at https://github.com/ljjcoder/EHTDI.
	\end{abstract}
	
	
	
	
	% \Description{Enjoying the baseball game from the third-base
		% seats. Ichiro Suzuki preparing to bat.}
	% \label{fig:teaser}
	
	\maketitle
	\begin{figure}
		\centering
		\includegraphics[width=0.95\linewidth]{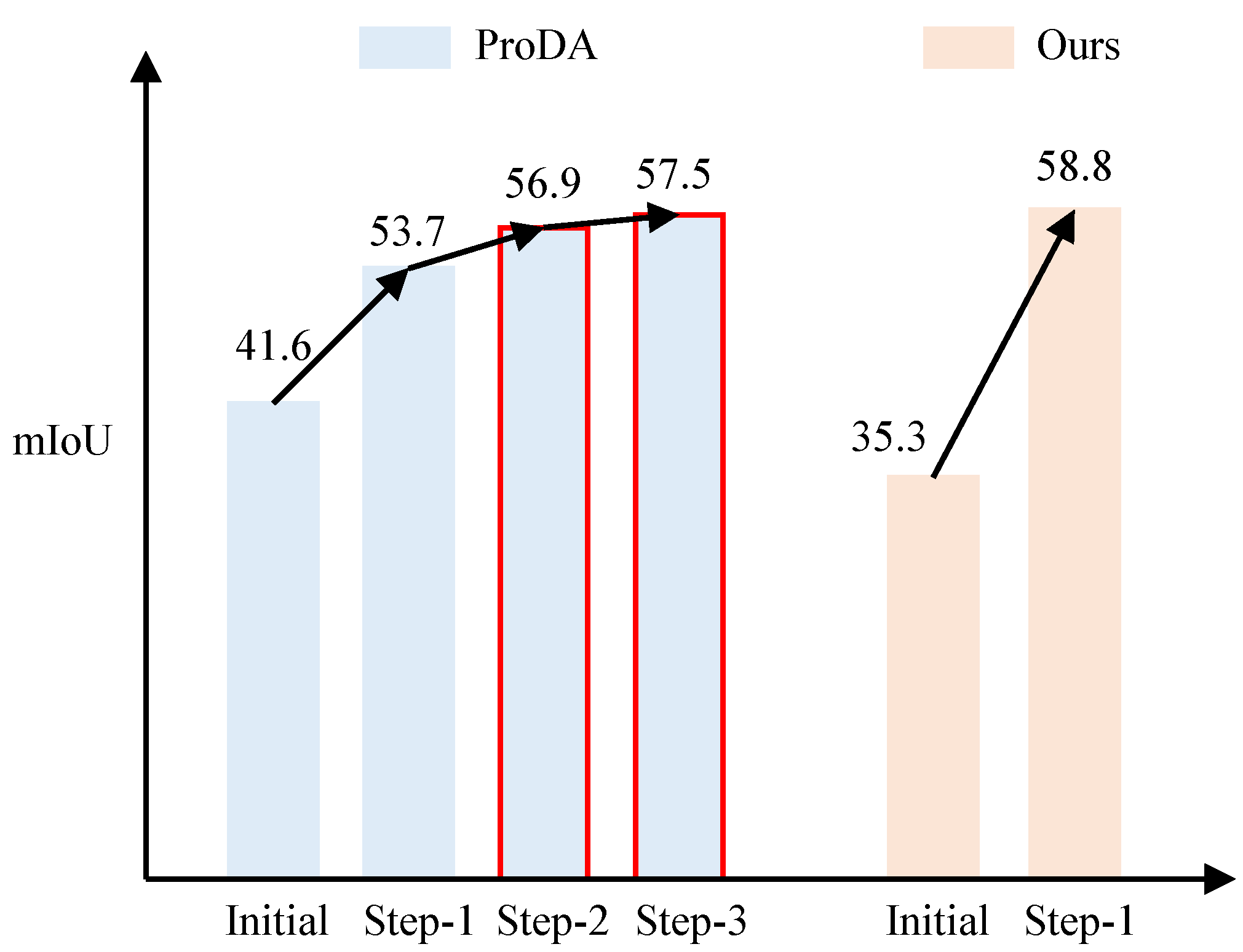}
		\vspace{-0.25cm}
		\caption{ProDA~\cite{zhang2021prototypical} v.s Ours on GTA5~\cite{richter2016playing}. ProDA needs to be initialized with the model of ASOS~\cite{tsai2018learning} and it requires an additional two-step distillation stage (red box). Compared to ProDA, our method only requires a source-only model for initialization and does not require the distillation stage.}
		\label{figure:ProDA_Ours}
		%  \vspace{-3pt}
	\end{figure}
	
	\section{Introduction}
	
	The goal of semantic segmentation is to assign a semantic class label to each pixel, which is an important problem in computer vision. In recent years, deep neural networks have shown significant advantages in semantic segmentation tasks~\cite{chen2017deeplab,chen2018encoder,liu2019auto,long2015fully,zhao2017pyramid}. However, such methods often require a large amount of manually annotated training data, and such pixel-level annotations are expensive and time-consuming. Therefore, a natural idea to overcome this bottleneck is to use synthetic data to train the model and then directly apply it in real-world scenarios. However, due to the huge gap between synthetic data and real data, applying models trained on synthetic data (source domain) to real data (target domain) often leads to a sharp drop in performance. In the field of computer vision, domain adaptive segmentation have emerged to solve this problem. In particular, this paper focuses on unsupervised domain adaptive (UDA) semantic segmentation.%Therefore, how to use part of the labeled data to assist us in learning more robust representations on the target data (unlabeled) has become a hot research topic. %Here we can simply divide this type of task into two categories, namely consistent semantic space representation learning (CLRL) and inconsistent semantic space representation learning (ISSRL).

	In UDA semantic segmentation, the methods using distillation techniques~\cite{zhang2021prototypical,zhang2021multiple,wang2021cross,huang2021category} generally outperform non-distilled methods~\cite{melas2021pixmatch,tranheden2021dacs,gao2021dsp,wang2021uncertainty}. For example, the first distillation-based method, ProDA~\cite{zhang2021prototypical}, is still a barrier that non-distillation methods cannot stride. Although the distillation technique has shown strong abilities to improve the generalization performance of the model, it interrupts the end-to-end training process and require some special training tricks. For example, ProDA requires three training stages and needs to be initialized with the ASOS~\cite{tsai2018learning} model in the first stage. In addition, in the distillation stage, ProDA also needs to be initialized with the model of SimCLRV2~\cite{chen2020big}. 
	However, contemporaneous non-distilled methods~\cite{melas2021pixmatch,tranheden2021dacs,gao2021dsp,wang2021uncertainty} only require one training stage and do not need special training strategies, although their performance is not as good as ProDA.
	Recently, few methods~\cite{zhang2021multiple,wang2021cross,huang2021category} can outperform ProDA with the same backbone (DeepLab-V2~\cite{chen2017deeplab}). However, they still need employ the distillation techniques and require some more sophisticated training techniques, further increasing the complexity of the methods. In pursuit of simplicity, we hope to design a non-distillation method that can achieve competitive performance to complicated distillation methods under the same condition with DeepLab-V2 as the backbone.

\begin{figure}
			\centering
			\includegraphics[width=0.95\linewidth]{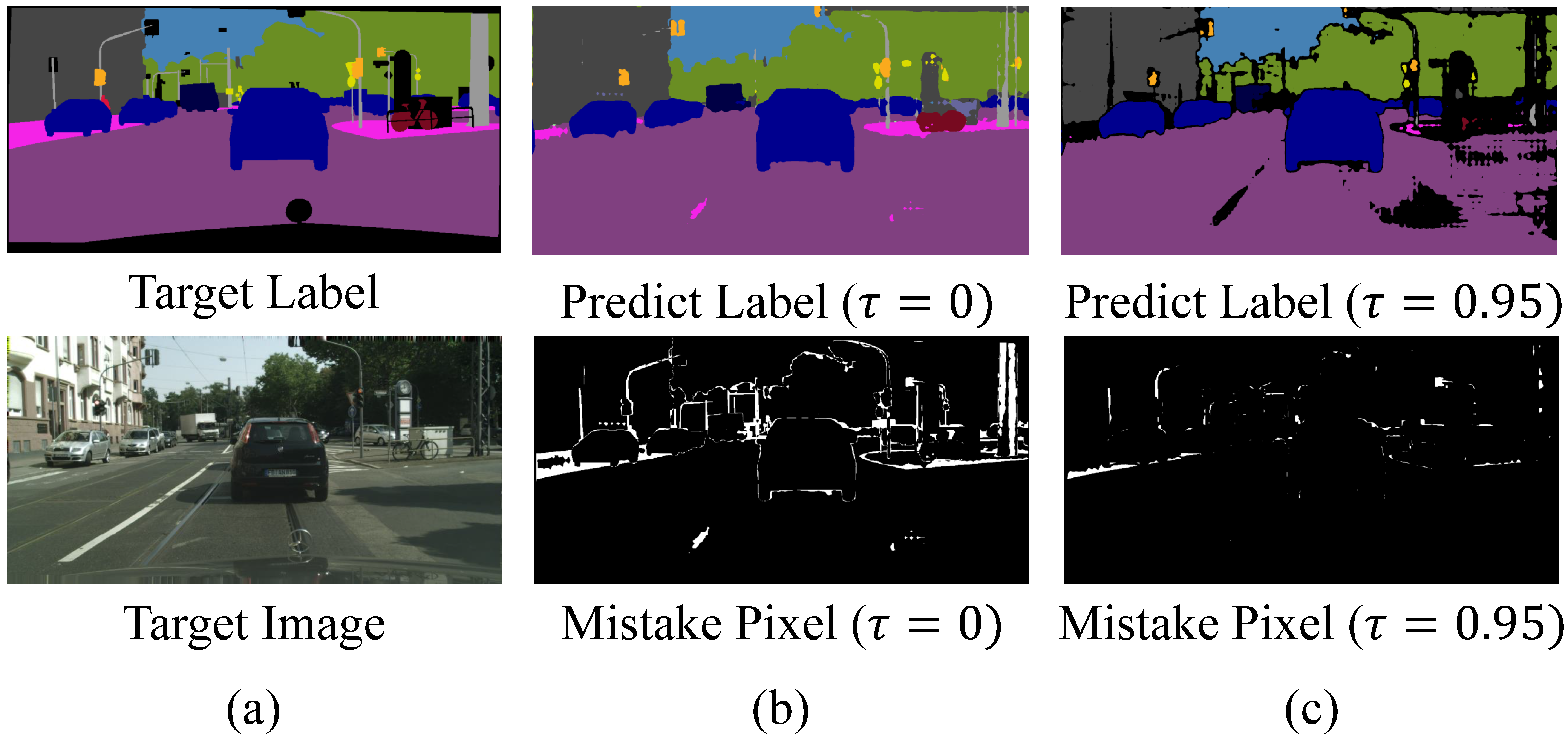}
			\vspace{-0.25cm}
			\caption{False foreground pixels under different thresholds $\tau$. White indicates wrong pixels. When the threshold $\tau$ is set to 0, most of the wrong pixels are concentrated near the boundary. When the threshold $\tau$ is set to 0.95, most of the wrong pixels are filtered out.}
			\label{figure:motivation_mistake}
			%  \vspace{-3pt}
		\end{figure}

	Besides the distillation-based methods, the methods employing pseudo-labeling techniques~\cite{tranheden2021dacs,melas2021pixmatch,zhou2021domain} have attracted many attention due to their simplicity and efficiency.
	These methods use the prediction results of the target image as labels, so that the target image can enjoy the same supervised training as the source image, thereby improving the performance of the model in the target image. %{\color{red}We know that the interior pixels in the target objects can easily obtain the correct pseudo-labels. Therefore, 
		With the enhancement of pseudo-labeling technology, the current method has good enough classification performance in the interior pixels of the target objects, and the bottleneck is mainly concentrated on the wrong prediction of edge pixels.
		As shown in Figure~\ref{figure:motivation_mistake}(b), wrong pixels are almost located in the boundary area, and the inner area of the object has few errors. Therefore, the key to improve segmentation performance is to improve the accuracy of the boundary pixels of the target images.
		However, since the pseudo-labels of the boundary pixels of the target domain are often wrong, it is inherently difficult for the pseudo-label techniques to improve the accuracy of boundary pixels in a supervised learning manner.
		
		%Through the above analysis, in order to improve the performance, we need to obtain the labels of the boundary pixels of the target domain. However, the difficulty now is that we cannot obtain boundary samples with accurate labels because the prediction of boundary pixels is always inaccurate. 
		%Although the edge pixels of the target image cannot get the correct labels, 
		In particular, we noticed that the prediction results of the interior pixels of objects in the target image are usually accurate. If we convert these interior pixels into boundary pixels, we would get the target domain boundary pixels with accurate labels. %To this end, we need to determine which pixels are interior. 
		Usually, the interior pixels have high confidence because there is no much ambiguity. As shown in Figure~\ref{figure:motivation_mistake}(c), when a high threshold is adopted, most of the remaining pixels are interior pixels with correct labels. Therefore, we choose to utilize the high-confidence target domain pixels to construct boundaries. % In order to effectively construct boundary pixels, 
		Here we particularly figure out the definition of the boundary. Generally speaking, the boundary refers to the area where the pixel-level semantics change (\textit{i.e.}, different categories are connected). Since there are multiple semantic categories around the pixels in these areas, the extracted features are usually mixed with other category information, leading to misjudgment. If the interior pixels of objects with correct labels (high-confidence pixels) in the target domain are placed in the regions with different semantics, we would obtain "new boundary" pixels with accurate labels. Then learning on these pixels can indirectly enhance the model to distinguish the boundary pixels of actual objects. %According to the above analysis, constructing high-quality target domain boundary pixels means  placing high-confidence target domain pixels at the locations of semantic change, which not only lie in the class intersection area but also have accurate labels. 

\begin{figure}
			\centering
			\includegraphics[width=0.95\linewidth]{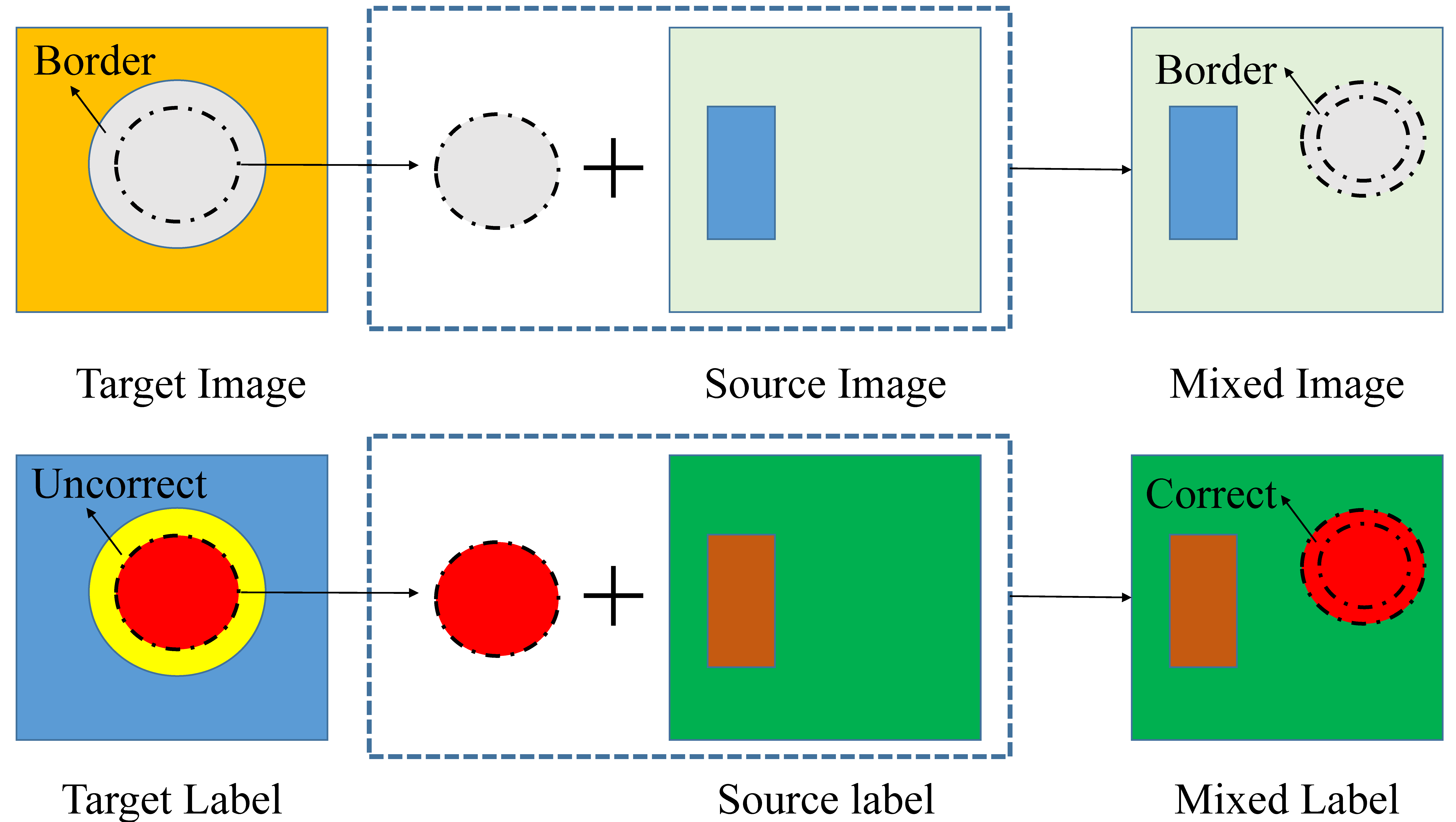}
			\vspace{-0.25cm}
			\caption{Illustration of the process of generating high-quality target domain boundary pixels. The boundary pixels of the target image are usually mispredicted (yellow region), but the inner pixels have the correct pseudo-labels (red region). Copies interior pixels into the source image so that some of the interior pixels become boundary pixels (dotted areas in the mixed image).}
			\label{figure:motivation}
			%  \vspace{-3pt}
		\end{figure}

		In this paper, we first propose a novel mix-up strategy to generate high-quality target domain boundaries with ground-truth label.
		Unlike DACS~\cite{tranheden2021dacs}, which explicitly learns cross-domain invariant source domain boundary representations by copying the parts of the source image to the target image, we directly reinforce the learning of the target boundary pixels by constructing target boundary pixels with correct labels.
		Our core idea is to copy the high-confidence pixels in the target image to the source image so that some high-confidence target pixels are in the position of category semantic change, \textit{i.e.}, becoming boundary pixels. Figure~\ref{figure:motivation} depicts the process of generating high-quality target domain boundary pixels. Furthermore, since the target domain lacks ground-truth supervision signals, the learned features in target domain are not discriminative enough. To address this issue, we propose a multi-level contrastive learning to explore more discriminative target domain representations.Specifically, we use both pixel-to-pixel and prototype-to-prototype contrastive learning to obtain more discriminative feature representations. By combining the proposed two techniques, our method successfully circumvents complex distillation techniques while achieving the state-of-the-art performance. As shown in Figure~\ref{figure:ProDA_Ours}, 
		our method can beat the distillation-based ProDA and does not require any special training tricks, such as ASOS~\cite{tsai2018learning} initialization.
		
		%%%
		Our main contributions are three-folds. 
		\begin{enumerate}
			\item We propose a novel high-confidence target domain pixel cross-domain mixing strategy (HTCM) to construct high-quality target domain boundary pixels to enhance the learning of target domain boundaries. 
			
			\item We propose multi-level contrastive learning to make target domain features more discriminative. Multi-level contrastive learning constrains the consistency from both pixel-level and prototype-level to improve the learning of target domain features.
			
			\item We construct an efficient non-distillation method. To the best of our knowledge, this is the first work that does not use the distillation technique while outperforming the distillation-based methods on both GTA5~\cite{richter2016playing} and SYNTHIA~\cite{ros2016synthia} datasets.
		\end{enumerate}
		
		\section{Related Work}
		\subsection{Unsupervised Domain Adaptive Semantic Segmentation}
		
		Early unsupervised domain adaptive (UDA) semantic segmentation methods~\cite{hoffman2018cycada,huang2020contextual,tsai2018learning,wang2020differential,yang2020fda,zhang2020joint}, usually use adversarial training to align source and target domain images.
		However, these methods often suffer from instability in the training process~\cite{chen2020adversarial,gulrajani2017improved}. Different from adversarial-based methods, there are methods utilizing pseudo-labels~\cite{tranheden2021dacs,melas2021pixmatch,zhou2021domain} and contrastive learning~\cite{zhou2021domain,xie2021spcl} to improve the performance. These methods are more concise than methods based on adversarial training.
		In addition to the above technical routes, ProDA~\cite{zhang2021prototypical} surpasses the above methods by using distillation technology. Subsequent works which outperform ProDA methods~\cite{zhang2021multiple,wang2021cross,huang2021category}, generally employ distillation techniques and use much more sophisticated training techniques. For example, MFA~\cite{zhang2021multiple} proposes multiple fusion adaptation to fuse three kinds of information and combine distillation technology to further improve the performance of ProDA. CRA~\cite{wang2021cross}, based on ProDA, aligns the distribution of confident pixels and unconfident pixels in the target domain through an additional cross-region adaptation operation.
		
		In this paper, we propose an efficient and non-distillation method that leverages well-designed pseudo-labeling techniques and contrastive learning techniques to reach or even surpass distillation-based methods.

		\subsection{Pseudo Labels}
		
		The workflow of the pseudo-label methods can be roughly divided into the following two steps: first, generating pseudo-labels in the unlabeled data, and second, finetuning the model on these pseudo-labels. Early methods~\citep{zou2018unsupervised, zou2019confidence,feng2020dmt,feng2020semi} were usually iteratively performed in an offline manner. Therefore, these methods often require multiple training stages and are very cumbersome.
		Recently, some works~\cite{tranheden2021dacs,melas2021pixmatch,zhou2021domain,xu2019self} have greatly simplified the training process through online pseudo-label techniques. For example, Pixmatch~\cite{melas2021pixmatch} uses random transformations to obtain a pair of strongly and weakly transformed images. Then the strongly transformed images can then enjoy the "supervised-form" of the training process with the pseudo-labels produced by the weaker transformations. DACS~\cite{tranheden2021dacs} proposes to construct a mixed domain by copying image patches from the source domain to the target domain and learn the consistency between the student model and the teacher model under different contexts. In this way, it explicitly learns the cross-domain consistency of source boundaries.
		The DSP~\cite{gao2021dsp} additionally introduces a mix up strategy between source domain images on the basis of DACS to further improve the performance. BAPA~\cite{liu2021bapa} increases the loss weight of boundary pixels generated by DACS to improve classification accuracy of boundary pixels. The above methods use the mix up strategy of DACS to strengthen the learning of boundary samples, but DACS only explicitly generates source domain boundaries and does not explicitly construct target domain boundaries, resulting in insufficient learning of the boundary pixels of the target domain.
		
		Different from these previous methods, we improve the performance of the model by explicitly constructing high-quality target domain boundary pixels.

		\subsection{Contrastive Learning}
		The goal of contrastive learning is to learn a model that can extract semantically compact features from raw images by encouraging positive pairs to be close and pulling negative pairs.
		Many works~\cite{chen2020simple, he2020momentum, grill2020bootstrap, caron2020unsupervised,zhang2022low,li2021semantic} have demonstrated the effectiveness of contrastive learning. In these works, an InfoNCE~\citep{oord2018representation} loss is usually adopted to implement contrastive learning.
		
		Some recent works introduce contrastive learning into UDA semantic segmentation.
		RCCR~\cite{zhou2021domain} builds different contexts for a target domain region through cutmix between the target domain and the source domain. Then they use pixel-to-pixel contrastive learning to maximize the difference of inter-region pixels and minimize the inconsistency of intra-region pixels. SPCL~\cite{xie2021spcl} propose a novel semantic prototype-based contrastive learning framework to achieve pixel and prototype alignment. 
		
		Different from these methods, 
		we design a multi-level contrastive learning loss, which both considers pixel-to-pixel and prototype-to-prototype contrastive learning losses, to explore more discriminative target domain feature representations. %Furthermore, to introduce semantic information (classifier weights), we compute the contrastive learning loss using post-classifier probabilities instead of pre-classifier features.

		\section{Preliminary}
		\begin{figure*}
			\centering
			\includegraphics[width=0.95\linewidth]{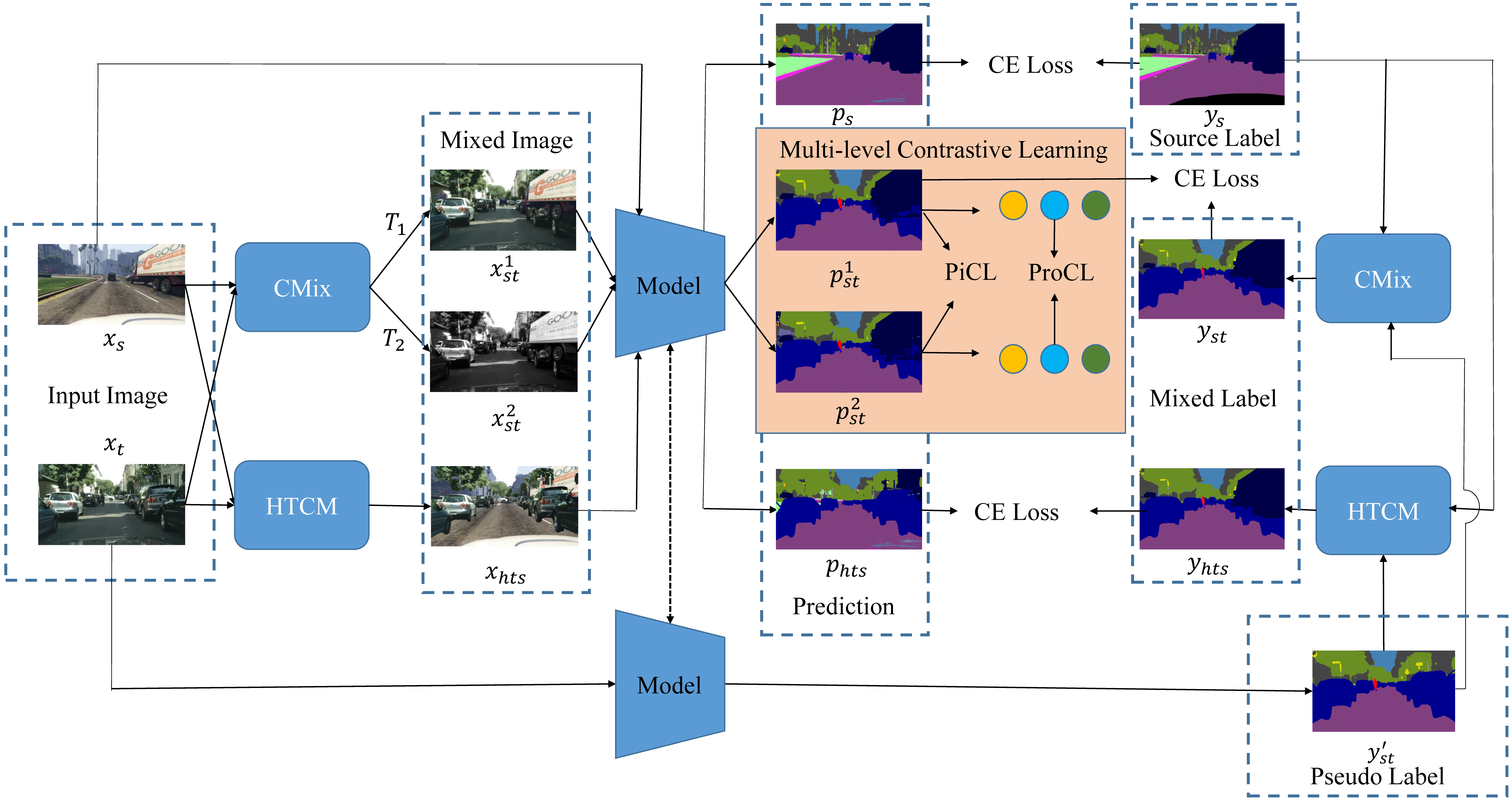}
			\caption{Framework of our method. CMix represents the class mix used in DACS. $T_1$ and $T_2$ represent two random transformations. In addition to Cmix, our method generates high-quality target domain boundary pixels via HTCM and performs multi-level contrastive learning on the mixed image $x_{st}$.}
			\label{figure:Framework}
			\vspace{-10pt}
		\end{figure*}
		Unsupervised domain adaptive (UDA) semantic segmentation contains two important datasets: an annotated source dataset $\mathbb{D}_S=\{(x_s,y_s)\}_{s=1}^{n_S}$ and an unlabeled target dataset $\mathbb{D}_T=\{x_t\}_{t=1}^{n_T}$. $\mathbb{D}_s$ and $\mathbb{D}_t$ share the same $C$ classes. The goal of UDA semantic segmentation is to use $\mathbb{D}_s$ and $\mathbb{D}_t$ to train a semantic segmentation model $M$ such that $M$ can correctly predict the class of each pixel in the target domain image. Generally, the model $M=G \circ 
		E$ can be regarded as a composite of a feature extractor $E$ and a classifier $G$. 
		
		To solve this problem, we can directly train a segmentation model with source data. Specifically, given a source image $x_s$ and its corresponding label $y_s$, the segmentation loss on the source domain can be defined as:
		\begin{equation}
			\mathbb{L}_{s} = -\sum_{i=1}^{H\times W}\mathbb{L}_{ce}(p_{s,i},y_{s,i}),
		\end{equation}
		where $p_{s,i}$ represents the predicted probability of pixel $x_{s,i}$. $\mathbb{L}_{ce}$ is the standard cross-entropy loss. Due to the domain shift in the source and target domains, it is difficult for the model trained in the above way to generalize well to the target domain data.
		To make the model better fit the target data, we can optimize the cross-entropy loss of the target image $x_t$ with the pseudo-label $y^{'}_t$. 
		In order to further improve the performance of the model on the target domain, we follow the mix-up strategy of DACS~\cite{tranheden2021dacs} to construct a mixed image $x_{st}$ and a mixed label $y_{st}$. Specifically, we randomly select half of the classes in the source image to generate a random mask $M_s$ and utilize $M_s$ to generate the mixed images from the source image and the target image.
		\begin{equation}
			x_{st} = x_s \odot M_s+x_t \odot (1-M_s).
		\end{equation}
		Similarly, the labels of the mix images can be obtained:
		\begin{equation}
			y_{st} = y_s \odot M_s+y^{'}_t \odot (1-M_s).
		\end{equation}
		Similar to the source domain segmentation loss, the segmentation loss of the mixed image is defined as follows:
		\begin{equation}
			\mathbb{L}_{st} = -\sum_{i=1}^{H\times W}\mathbb{L}_{ce}(p_{st,i},y_{st,i}).
		\end{equation}

		\section{Method}
		\label{method}
		In this work, we aim to build an efficient unsupervised domain adaptive semantic segmentation method that does not require distillation techniques. To achieve our goal, we propose two novel self-supervised learning methods. 
		Figure~\ref{figure:Framework}
		illustrates the overall architecture of our proposed method.
		
		In particular, we propose a novel high-confidence target domain pixel cross-domain mixing strategy (HTCM) to construct target domain boundary pixels with correct semantic labels. In principle, HTCM only copies the high-confidence pixels in the target domain to the source domain image so that some high-confidence target domain pixels become boundary pixels. In addition, we design a multi-level contrastive learning to improve the discriminativeness of features from both pixel-level and prototype-level.
		\subsection{Generating High-quality Target Boundary Pixels}
		
		In general, only the inner pixels of the target image can obtain the correct pseudo-labels, while it is difficult to obtain accurate labels for the boundary pixels of the target image. As a result, when training target domain images directly with pseudo-labels, only the inner pixels can be effectively trained with supervision, while the boundary pixels of the target image are difficult to be effectively trained. To alleviate this problem, we need to obtain target boundaries with correct labels, strengthening the learning of target boundary pixels.

		\begin{figure}
			\centering
			\includegraphics[width=0.95\linewidth]{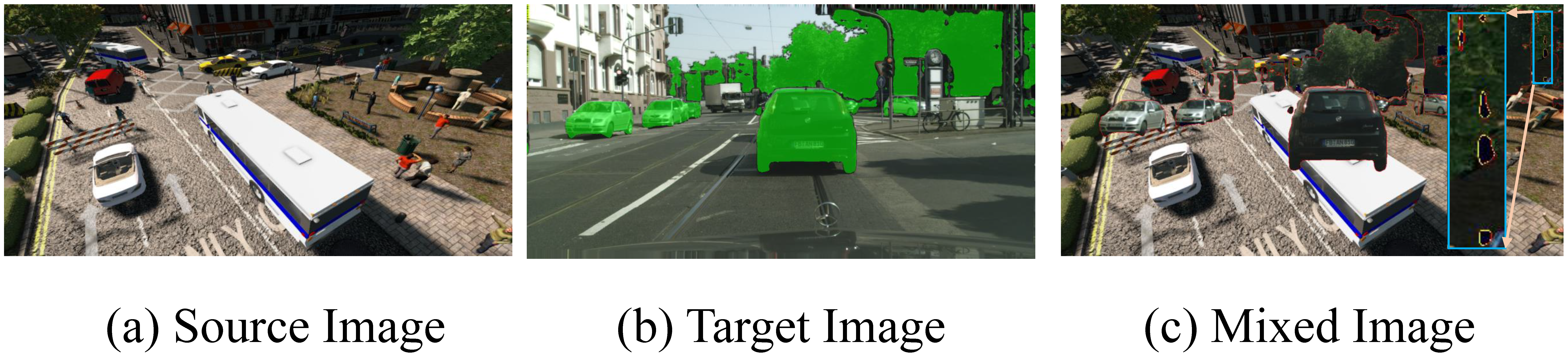}
			\vspace{-0.25cm}
			\caption{Visual demonstration of the HTCM algorithm. The green parts represent high-confidence target domain pixels. The red part of the mixed image represents the border pixels with the correct label, while the yellow part represents the border pixels with the wrong label. It can be seen that most of the border pixels in the mixed image have the correct labels.}
			\label{figure:method_GHTB}
			%  \vspace{-3pt}
		\end{figure}
		In this section, we propose a simple yet effective high-confidence target domain pixel cross-domain mixing strategy (HTCM) to generate high-quality target domain boundary pixels. 
		Generating high-quality target boundary pixels means constructing target pixels located in semantic mutation regions with correct pseudo-labels. 
		In the pseudo-labels of the target domain pixels, the labels of high-confidence pixels tend to be correct, but these pixels are usually located inside objects rather than boundaries. We want to have these correctly labeled target domain pixels at boundary locations so that we can get the target domain boundary pixels with the correct labels. To do this, we paste these pixels into a source image so that some of the target domain pixels are at the border. Specifically, for a target domain image $x_t$, we can get its corresponding predicted probability map $p_
		t^{i}$. Then we get a high-confidence template $M_{ht}$, which is the selected indicator for the pixels whose predicted probability is greater than a threshold $\tau$. Finally, we use the obtained templates $M_{ht}$ to construct high-quality boundary samples as follows:
		\begin{equation}
			x_{hts} = x_s \odot (1-M_{ht})+x_t \odot M_{ht}.
		\end{equation}
		Similarly, the labels of the mix images can be obtained:
		\begin{equation}
			y_{hts} = y_s \odot (1-M_{ht})+\hat{y}_t \odot M_{ht}.
		\end{equation}
		
		The segmentation loss of the mixed image $x_{hts}$ is defined as follows:
		\begin{equation}
			\mathbb{L}_{hts} = -\sum_{i=1}^{H\times W}\mathbb{L}_{ce}(p_{hts,i},y_{hts,i}).
		\end{equation}
		\subsection{Multi-level Contrastive Learning}
		\label{mcl}
		Due to the lack of explicit supervision signals in the target domain data, the features of the target domain are not sufficiently discriminative. Fortunately, contrastive learning can improve the distinguishability of unlabeled data~\cite{chen2020simple, he2020momentum, grill2020bootstrap, caron2020unsupervised}, so we employ contrastive learning to further improve our model.
		%Since the contrastive learning technique shows strong generalization ability in unsupervised tasks~\cite{chen2020simple, he2020momentum, grill2020bootstrap, caron2020unsupervised}, we further improve the transferability of the model through contrastive learning. 
		In particular, we design a multi-level contrastive learning loss to improve feature distinguishability from both prototype-level and pixel-level. Since we want the features to be invariant across domains, we choose to perform contrastive learning on the mixed images instead of directly on the target image. Because the prototype of the mixed image will contain the information of the source domain and the target domain, the features of the same category in the source domain and the target domain will be as consistent as possible during prototype-level 
		contrastive learning. At the same time, because the target pixels in the mixed image $x_{hts}$ are all high-confidence samples, they have the correct pseudo-labels. 
		It means that all the pixels of the mixed image $x_{hts}$ can be trained like the source domain image. At this time, using contrastive learning to improve the discriminativeness of features will not have much effect.  Therefore, we choose to perform contrastive learning on mixed images $x_{st}$.

		Specifically, for the mixed image $x_{st}$, we first obtain the transformed mixed images $x^{1}_{st}$ and $x^{2}_{st}$ through two different random transformations. Then, we use the feature extractor $E$ to obtain the feature maps $\mathbf{f}^{1}_{st}$, $\mathbf{f}^{2}_{st}$ corresponding to $x^{1}_{st}$ and $x^{2}_{st}$, whose resolution $H^{'} \times W^{'}$ is $\frac{1}{8}$ of the native resolution $H \times W$.  In order to alleviate the interference of wrong samples on the prototype, we only use the features with high confidence to calculate the prototype. Therefore, we can get the prototype of each category as follows.
		
		\begin{equation}
			p^{1,c}_{st} =-\frac{1}{\sum M^{c}_{st}}\sum \mathbf{f}^{1}_{st} \odot (1-M^{c}_{st}),
		\end{equation}
		\begin{equation}
			p^{2,c}_{st} =-\frac{1}{\sum M^{c}_{st}}\sum \mathbf{f}^{2}_{st} \odot (1-M^{c}_{st}).
		\end{equation}
		
		$M^{c}_{st}$ is a binary mask, $1$ indicates that the predicted category of this position is $c$ and the predicted probability is greater than $\tau$, 0 indicates that the predicted category is other categories or predicted probability is less than $\tau$. For a query sample $\mathbf{p}^{1,c}_{st}$, the prototype $\mathbf{p}^{2,c}_{st}$ is the positive and all other prototype are the negative. In domain adaptation tasks, it is not enough to improve the discriminativeness of features, and we also need to narrow the distance between features and classifier weights~\cite{li2021semantic}. Therefore, following PCL~\cite{li2021semantic}, we compute the contrastive loss using the probabilities instead of features and then the prototype-level contrastive loss (ProCL) is defined as:%In order to strengthen the consistency of features and class weights, we do not use the prototype directly but use the prototype output by the classifier $
		%G$ to calculate the prototype-level contrastive loss (ProCL). 
		
		\begin{align}
			\ell_{\mathbf{p}^{1,c}_{st}}\!=\!
			-\!
			\log \frac{ \exp(s_1H(\mathbf{p}^{1,c}_{st} ,\tilde{\mathbf{p}}^{2,c}_{st}) }{ \sum_{j \neq c} \exp( s_1H(\mathbf{p}^{1,c}_{st},\mathbf{p}^{1,j}_{st})\!+\!\sum_{k} \exp( s_1H(\mathbf{p}^{1,c}_{st},\tilde{\mathbf{p}}^{2,k}_{st})}.
			\label{eq:FCL1}
		\end{align}
		
		In addition to prototype-level contrastive learning, we further consider pixel-level contrastive learning. For a query sample $\mathbf{f}^{1,i}_{st}$, the feature $\mathbf{f}^{2,i}_{st}$ at the corresponding position of another transformed image is taken as a positive sample, and all the other features are taken as negative samples. The pixel-level contrastive learning loss (PiCL) is defined as follows:
		\begin{align}
			\ell_{\mathbf{f}^{1,i}_{st}}\!=\!
			-\!
			\log \frac{ \exp(s_2H(\mathbf{f}^{1,i}_{st}, \mathbf{f}^{2,i}_{st}) }{ \sum_{j \neq i} \exp( s_2H(\mathbf{f}^{1,i}_{st}, \mathbf{f}^{1,j}_{st})\!+\! \sum_{k} \exp( s_2H(\mathbf{f}^{1,i}_{st},\mathbf{f}^{2,k}_{st})}.
			\label{eq:FCL}
		\end{align}
		where $s_1$ and  $s_2$ are the scaling factors, and we set $s_1$ to 7 and set $s_2$ to 20. $H(x,y)=\sigma(G(x))^{\top}\sigma(G(y))$  calculates the inner product of the classification probabilities corresponding to the two vectors and $\sigma$ represents the softmax function.

		\begin{figure*}
			\centering
			\includegraphics[width=0.95\linewidth]{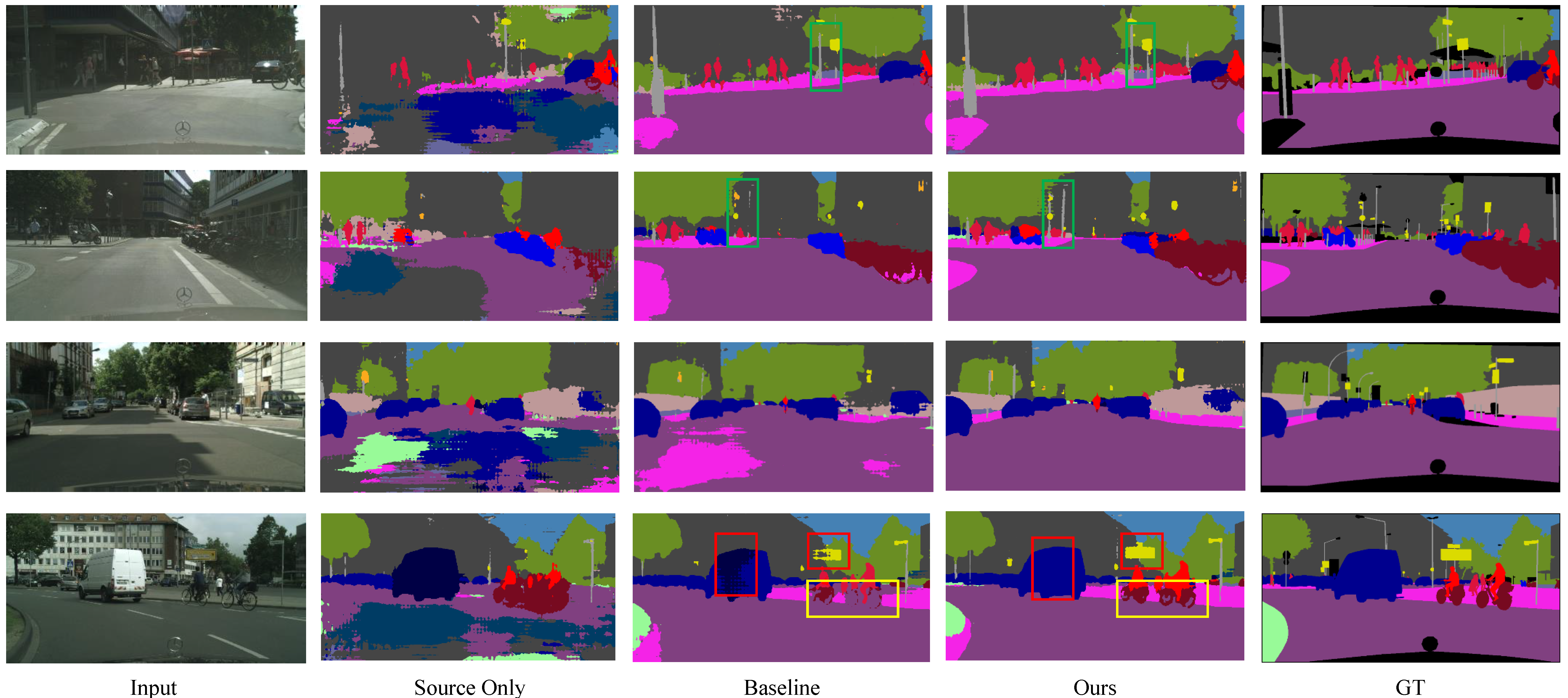}
			\caption{Visualization of the segmentation results on GTA5$\rightarrow$Cityscapes task. Baseline means that the model is only trained with $\mathbb{L}_{s}$ and $\mathbb{L}_{st}$.}
			\label{figure:vis}
			\vspace{-10pt}
		\end{figure*}
		\subsection{Loss Function}
		Our loss function is defined as:
		\begin{align}
			L =\mathbb{L}_{s} +\lambda_1 (\mathbb{L}_{st}+\mathbb{L}_{hts} ) +\lambda_2 (\mathbb{L}_{pro}+\mathbb{L}_{pixel} ).
			\label{loss}
		\end{align}
		Here $\mathbb{L}_{pro}=\sum_{c} (\ell_{\mathbf{p}^{1,c}_{st}}+\ell_{\mathbf{p}^{2,c}_{st}})$ and $\mathbb{L}_{pixel}=\sum_{i} (\ell_{\mathbf{f}^{1,i}_{st}}+\ell_{\mathbf{f}^{2,i}_{st}})$. $\lambda_1$ and $\lambda_2$ are the hyper-parameters to balance different losses.

		\begin{table*}[h]
			\fontsize{8}{9}\selectfont
			\setlength{\tabcolsep}{0.30em}
			\def\arraystretch{1.15}
			
			\begin{center}
				\caption{Results on the GTA5$\rightarrow$Cityscapes benchmark. D means using distillation technique.}
				\addtolength{\leftskip} {-0.5cm} % increase (absolute) value if needed
				% \addtolength{\rightskip}{-1.8cm}

				\vspace{1mm}
				\begin{tabular}{lll|ccccccccccccccccccc|c}
					
					& Method & Venue & \rotatebox{90}{road} & \rotatebox{90}{sdwk} & \rotatebox{90}{bld} & \rotatebox{90}{wall} & \rotatebox{90}{fnc} & \rotatebox{90}{pole} & \rotatebox{90}{lght} & \rotatebox{90}{sign} & \rotatebox{90}{veg.} & \rotatebox{90}{trrn.} & \rotatebox{90}{sky} & \rotatebox{90}{pers} & \rotatebox{90}{rdr} & \rotatebox{90}{car} & \rotatebox{90}{trck} & \rotatebox{90}{bus} & \rotatebox{90}{trn} & \rotatebox{90}{mtr} & \rotatebox{90}{bike} & mIoU \\ \toprule    
					\midrule 
					& Source only	&  & 58.6 & 13.5 & 64.1 & 16.8 & 14.0 & 23.8 & 36.2 & 19.8 & 80.3 & 19.5 & 66.3 & 58.9 & 28.7 & 64.9 & 28.7 & 3.5 & 8.0 & 29.6 & 35.3 & 35.3 \\ 
					\midrule
					
					%& CLST~\cite{marsden2021contrastive} & arXiv 2021 & 92.8  & 53.5 & 86.1 & 39.1 & 28.1 & 28.9 & 43.6 & 39.4 & 84.6 & 35.7 & 88.1 & 63.9 & 38.3 & 86.0 & 41.6 & 50.6 & 0.1 & 30.4 & 51.7 & 51.6 \\
					
					&  DACS~\cite{tranheden2021dacs} & WACV 2020  & 89.9  & 39.7 & 87.9 & 30.7 & 39.5 & 38.5 & 46.4 & 52.8 & 88.0 & 44.0 & 88.8 & 67.2 & 35.8 & 84.5 & 45.7 & 50.19 & 0.0 & 27.3 & 34.0 & 52.1 \\
					
					& SPCL~\cite{xie2021spcl} & arXiv 2021  & 90.3  & 50.3 & 85.7 & 45.3 & 28.4 & 36.8 & 42.2 & 22.3 & 85.1 & 43.6 & 87.2 & 62.8 & 39.0 & 87.8 & 41.3 & 53.9 & 17.7 & 35.9 & 33.8 & 52.1 \\
					
					& RCCR~\cite{zhou2021domain} & arXiv 2021  & 93.7  & 60.4 & 86.5 & 41.1 & 32.0 & 37.3 & 38.7 & 38.6 & 87.2 & 43.0 & 85.5 & 65.4 & 35.1 & 88.3 & 41.8 & 51.6 & 0.0 & 38.0 & 52.1 & 53.5 \\
					
					&  SAC~\cite{araslanov2021self} & CVPR 2021  & 90.4 & 53.9 & 86.6 & 42.4 & 27.3 & 45.1 & 48.5 & 42.7 & 87.4 & 40.1 & 86.1 & 67.5 & 29.7 & 88.5 & 49.1 & 54.6 & 9.8 & 26.6 & 45.3 & 53.8 \\ 
					%& MetaCorrection~\cite{guo2021metacorrection} & CVPR 2021  & 92.8  & 58.1 & 86.2 & 39.7 & 33.1 & 36.3 & 42.0 & 38.6 & 85.5 & 37.8 & 87.6 & 62.8 & 31.7 & 84.8 & 35.7 & 50.3 & 2.0 & 36.8 & 48.0 & 52.1 \\
					& Pixmatch~\cite{melas2021pixmatch} & CVPR 2021  & 91.6  & 51.2 & 84.7 & 37.3 & 29.1 & 24.6 & 31.3 & 37.2 & 86.5 & 44.3 & 85.3 & 62.8 & 22.6 & 87.6 & 38.9 & 52.3 & 0.7 & 37.2 & 50.0 & 50.3 \\
					& DPL~\cite{cheng2021dual} & ICCV 2021  & 92.8  & 54.4 & 86.2 & 41.6 & 32.7 & 36.4 & 49.0 & 34.0 & 85.8 & 41.3 & 83.0 & 63.2 & 34.2 & 87.2 & 39.3 & 44.5 & 18.7 & 42.6 & 43.1 & 53.3 \\
					& BAPA~\cite{liu2021bapa} & ICCV 2021 & 94.4  & 61.0 & 88.0 & 26.8 & 39.9 & 38.3 & 46.1 & 55.3 & 87.8 & 46.1 & 89.4 & 68.8 & 40.0 & 90.2 & 60.4 & 59.0 & 0.00 & 45.1 & 54.2 & 57.4 \\
					& UPST~\cite{wang2021uncertainty} & ICCV 2021 & 90.5  & 38.7 & 86.5 & 41.1 & 32.9 & 40.5 & 48.2 & 42.1 & 86.5 & 36.8 & 84.2 & 64.5 & 38.1 & 87.2 & 34.8 & 50.4 & 0.2 & 41.8 & 54.6 & 52.6 \\
					& DSP~\cite{gao2021dsp} & ACM MM 2021  & 92.4  & 48.0 & 87.4 & 33.4 & 35.1 & 36.4 & 41.6 & 46.0 & 87.7 & 43.2 & 89.8 & 66.6 & 32.1 & 89.9 & 57.0 & 56.1 & 0.0 & 44.1 & 57.8 & 55.0 \\

					\midrule
					%\multicolumn{23}{c}{With Knowledge Distillation} \\ 
					\midrule 
					&  MFA  (D)~\cite{zhang2021multiple}  & BMVC 2021 & 93.5 & 61.6 & 87.0 & 49.1 & 41.3 & 46.1 & 53.5 & 53.9 & 88.2 & 42.1 & 85.8 & 71.5 & 37.9 & 88.8 & 40.1 & 54.7 & 0.0 & 48.2 & 62.8 & 58.2 \\
					
					&  ProDA  (D)~\cite{zhang2021prototypical} & CVPR 2021  & 87.8 & 56.0 & 79.7 & 46.3 & 44.8 & 45.6 & 53.5 & 53.5 & 88.6 & 45.2 & 82.1 & 70.7 & 39.2 & 88.8 & 45.5 & 59.4 & 1.0 & 48.9 & 56.4 & 57.5 \\
					
					& ProDA + CaCo  (D)~\cite{huang2021category} & CVPR 2022  & 93.8 & 64.1 & 85.7 & 43.7 & 42.2 & 46.1 & 50.1 & 54.0 & 88.7 & 47.0 & 86.5 & 68.1 & 2.9 & 88.0 & 43.4 & 60.1 & 31.5 & 46.1 & 60.9 & 58.0 \\
					
					& ProDA + CRA  (D)~\cite{wang2021cross} & arXiv 2021  & 89.4 & 60.0 & 81.0 & 49.2 & 44.8 & 45.5 & 53.6 & 55.0 & 89.4 & 51.9 & 85.6 & 72.3 & 40.8 & 88.5 & 44.3 & 53.4 & 0.0 & 51.7 & 57.9 &  58.6 \\

					\midrule \midrule 
					& Ours & & 95.4 & 68.8 & 88.1 & 37.1 & 41.4 & 42.5 & 45.7 & 60.4 & 87.3 & 42.6 & 86.8 & 67.4 & 38.6 & 90.5 & 66.7 & 61.4 & 0.3 & 39.4 & 56.1 & \bf 58.8 \\

					\bottomrule
				\end{tabular}
				\vspace{2mm}
				
				\label{table:gta}
			\end{center}
		\end{table*}

		% \end{landscape}
	% \clearpage 
	% \twocolumn

	% \clearpage
	\begin{table*}[h]
		\fontsize{8}{9}\selectfont
		\setlength{\tabcolsep}{0.30em}
		\def\arraystretch{1.15}
		\begin{center}
			\caption{Results on the SYNTHIA$\rightarrow$Cityscapes benchmark. mIoU-16 and mIoU-13 refer to mean intersection-over-union on the standard sets of 16 and 13 classes, respectively. Classes not evaluated are replaced by '-'. D means using distillation technique.}
			\addtolength{\leftskip} {-1.3cm} % increase (absolute) value if needed
			\addtolength{\rightskip}{-1.8cm}
			% \rowcolors{1}{white}{gray!15}
			
			\vspace{1mm}
			\begin{tabular}{lll|cccccccccccccccc|c|c}
				& Method & Venue  & \rotatebox{90}{road} & \rotatebox{90}{sdwk} & \rotatebox{90}{bld} & \rotatebox{90}{wall$^{*}$} & \rotatebox{90}{fnc$^{*}$} & \rotatebox{90}{pole$^{*}$} & \rotatebox{90}{light} & \rotatebox{90}{sign} & \rotatebox{90}{veg.} & \rotatebox{90}{sky} & \rotatebox{90}{pers} & \rotatebox{90}{rdr} & \rotatebox{90}{car} & \rotatebox{90}{bus} & \rotatebox{90}{mtr} & \rotatebox{90}{bike} & mIoU-16 & mIoU-13 \\ \toprule
				\midrule 
				& Source only	&  & 40.1 & 18.3 & 64.9 & 5.9 & 0.1 & 24.6 & 5.9 & 9.0 & 74.8 & 81.6 & 58.7 & 16.8 & 43.9 & 11.8 & 6.4 & 24.11 & 30.4 & 35.1\\
				\midrule 
				%&  AdaptSegNet~\cite{} & CVPR 2018 &	 79.2 &  37.2 &  78.8 &  10.5 &  0.3  &  25.1 &  9.9 &  10.5 &  78.2 &  80.5 &  53.5 &  19.6 &  67.0 &  29.5 &  21.6 &  31.3 &  39.5 &  45.9 \\ 
				
				%&  AdvEnt~\cite{} & CVPR 2019 &  87.0 &  44.1 &  79.7 &  9.6  &  0.6 &  24.3 &  4.8  &  7.2  &  80.1 &  83.6 &  56.4 &  23.7 &  72.7 &  32.6 &  12.8 &  33.7   &  40.8 &  47.6 \\
				
				%&  MaxSquare~\cite{chen2019domain} & ICCV 2019  &  82.9 &  40.7 &  80.3 &  10.2 &  0.8  &  25.8 &  12.8 &  18.2 &  82.5 &  82.2 &  53.1 &  18.0 &  79.0 &  31.4 &  10.4 &  35.6 &  41.4    &  48.2\\
				
				%&  CRST (MRENT)~\cite{} & ICCV 2019 &  69.6 &  32.6 &  75.8 &  12.2 &  1.8  &  35.3 &  23.3 &  29.5 &  77.7 &  78.9 &  60.0 &  28.5 &  81.5 &  25.9 &  19.6 &  41.8 &  43.4    &  49.6 \\
				
				%&  CRST (MRKLD)~\cite{} & ICCV 2019  &  67.7 &  32.2 &  73.9 &  10.7 &  1.6  &  37.4 &  22.2 &  31.2 &  80.8 &  80.5 &  60.8 &  29.1 &  82.8 &  25.0 &  19.4 &  45.3 &  43.8    &  50.1 \\
				
				%&  FADA~\cite{wang2020classes} & ECCV 2020  &  84.5 &  40.1 &  83.1 &  4.8  &  0.0 &  34.3 &  20.1 &  27.2 &  84.8 &  84.0 &  53.5 &  22.6 &  85.4 &  43.7 &  26.8 &  27.8   &  45.2 &  52.5 \\
				
				%&  FDA~\cite{yang2020fda} & CVPR 2020 &  79.3 &  35.0 &  73.2 &  -  &  -  &  -  & 19.9 & 24.0 &  61.7 &  82.6 &  61.4 &  31.1 &  83.9 &  40.8 &  38.4 &  51.1 & - &  52.5 \\

				%&  CLST~\cite{marsden2021contrastive} & arXiv 2021 & 88.0 & 49.2 & 82.2 & 16.3 & 0.4 & 29.2 & 31.8 & 23.9 & 84.1 & 88.0 & 59.1 & 27.2 & 85.5 & 46.6 & 28.9 & 56.5 & 49.8 & 57.8 \\
				&  DACS~\cite{tranheden2021dacs}  & WACV 2020 & 80.6 & 25.1 & 81.9 & 21.5 & 2.9 & 37.2 & 22.7 & 24.0 & 83.7 & 90.8 & 67.6 & 38.3 & 82.9 & 38.9 & 28.5 & 47.6 & 48.3 & 54.8 \\ 
				&  SPCL~\cite{xie2021spcl} & arXiv 2021 & 86.9 & 43.2 & 81.6 & 16.2 & 0.2 & 31.4 & 12.7 & 12.1 & 83.1 & 78.8 & 63.2 & 23.7 & 86.9 & 56.1 & 33.8 & 45.7 & 47.2 & 54.4 \\
				&  RCCR~\cite{zhou2021domain}  & arXiv 2021 &  79.4 & 45.3 & 83.3 & - & - & - & 24.7 & 29.6 & 68.9 & 87.5 & 63.1 & 33.8 & 87.0 & 51.0 & 32.1 & 52.1 & - & 56.8 \\

				&  SAC~\cite{araslanov2021self}  & CVPR 2021   &  89.3 & 47.2 & 85.5 & 26.5 & 1.3 & 43.0 & 45.5 & 32.0 & 87.1 & 89.3 & 63.6 & 25.4 & 86.9 & 35.6 & 30.4 & 53.0 & 52.6 & 59.3 \\
				%&  MetaCorrection~\cite{guo2021metacorrection} & CVPR 2021 & 92.6 & 52.7 & 81.3 & 8.9 & 2.4 & 28.1 & 13.0 & 7.3 & 83.5 & 85.0 & 60.1 & 19.7 & 84.8 & 37.2 & 21.5 & 43.9 & 45.1 & 52.5 \\
				&  Pixmatch~\cite{melas2021pixmatch} & CVPR 2021 & 92.5 & 54.6 & 79.8 & 4.8 & 0.1 & 24.1 & 22.8 & 17.8 & 79.4 & 76.5 & 60.8 & 24.7 & 85.7 & 33.5 & 26.4 & 54.4 & 46.1 & 54.5 \\
				& DPL~\cite{cheng2021dual} & ICCV 2021 & 87.5 & 45.7 & 82.8 & 13.3 & 0.6 & 33.2 & 22.0 & 20.1 & 83.1 & 86.0 & 56.6 & 21.9 & 83.1 & 40.3 & 29.8 & 45.7 & 47.0 & 54.2 \\
				& BAPA~\cite{liu2021bapa} & ICCV 2021 & 91.7 & 53.8 & 83.9 & 22.4 & 0.8 & 34.9 & 30.5 & 42.8 & 86.6 & 88.2 & 66.0 & 34.1 & 86.6 & 51.3 & 29.4 & 50.5 & 53.3 & 61.2 \\
				& UPST~\cite{wang2021uncertainty} & ICCV 2021 & 79.4 & 34.6 & 83.5 & 19.3 & 2.8 & 35.3 & 32.1 & 26.9 & 78.8 & 79.6 & 66.6 & 30.3 & 86.1 & 36.6 & 19.5 & 56.9 & 48.0 & 54.6 \\

				& DSP~\cite{gao2021dsp}  & ACM MM 2021  & 86.4 & 42.0 & 82.0 & 2.1 & 1.8 & 34.0 & 31.6 & 33.2 & 87.2 & 88.5 & 64.1 & 31.9 & 83.8 & 65.4 & 28.8 & 54.0 & 51.0 & 59.9 \\

				\midrule 
				\midrule 
				&  MFA  (D)~\cite{zhang2021multiple} & BMVC 2021  & 81.8 & 40.2 & 85.3 & - & - & - & 38.0 & 33.9 & 82.3 & 82.0 & 73.7 & 41.1 & 87.8 & 56.6 & 46.3 & 63.8 & - & 62.5 \\
				
				& ProDA  (D)~\cite{zhang2021prototypical}  & CVPR 2021 & 87.8 & 45.7 & 84.6 & 37.1 & 0.6 & 44.0 & 54.6 & 37.0 & 88.1 & 84.4 & 74.2 & 24.3 & 88.2 & 51.1 & 40.5 & 45.6 & 55.5 & 62.0 \\
				
				& ProDA + CRA  (D)~\cite{wang2021cross} & arXiv 2021  & 85.6 & 44.2 & 82.7 & 38.6 & 0.4 & 43.5 & 55.9 & 42.8 & 87.4 & 85.8 & 75.8 & 27.4 & 89.1 & 54.8 & 46.6 & 49.8 & 56.9 & 63.7 \\
				
				\midrule \midrule 
				& Ours  &  & 93.0 & 69.8 & 84.0 & 36.6 & 9.1 & 39.7 & 42.2 & 43.8 & 88.2 & 88.1 & 68.3 & 29.0 & 85.5 & 54.1 & 37.1 & 56.3 & \bf 57.8 & \bf 64.6 \\
				
				\bottomrule
			\end{tabular}
			\vspace{8mm}
			\label{table:synthia}
		\end{center}  
	\end{table*}
	
	\section{Experiments}
	\subsection{Experimental Setup}
	\subsubsection{Dataset}
	
	We evaluate the performance of our methods on the two standard UDA semantic segmentation tasks: GTA5$\rightarrow$Cityscapes and SYNTHIA$\rightarrow$Cityscapes. 
	\textbf{GTA5}~\cite{richter2016playing} is a synthetic dataset consisting of $24,966$ annotated images with $1914\times1052$ resolution taken from the GTA5 game, which is used as a source domain dataset. 
	\textbf{SYNTHIA}~\cite{ros2016synthia} is also a synthetic dataset consisting of $9,400$ annotated images with $1280\times720$ resolution, which is used as another source domain dataset.
	\textbf{Cityscapes}~\cite{cordts2016cityscapes} is a representative dataset in semantic segmentation and autonomous driving domain, which contains 5000 pixel-level annotated images with $2048\times1024$ resolution taken from real urban street scenes and is use as the target domain dataset. In all experiments, we can obtain $2975$ unlabeled target domain images for training and we use $500$ validation images for test.
	Following existing works~\cite{tranheden2021dacs,melas2021pixmatch}, we evaluate our method on 19 common categories for GTA5$\rightarrow$Cityscapes and both 16 and 13 common categories for SYNTHIA$\rightarrow$Cityscapes, respectively.

	\subsubsection{Implementation Details}
	
	Following common UDA protocols~\cite{tranheden2021dacs,melas2021pixmatch,zhou2021domain}, we use the DeepLab-v2 segmentation model~\cite{chen2017deeplab} with a ResNet-101~\cite{he2016deep} backbone pre-trained on ImageNet~\cite{deng2009imagenet} as our model. Following~\cite{chen2019domain}, we first perform warm-up training on source domain to obtain an initialized model. Then we train the model with our method. 
	Specifically, we use the ''poly'' learning weight decay and the power is set to 0.9. The SGD optimizer is implemented with weight decay $5\times10^{-4}$ and momentum 0.9. The learning rate is set at $2.5\times10^{-4}$ for backbone parameters and $2.5\times10^{-3}$ for others. The maximum iteration number is 250$k$ and we use an early stopping strategy. For the hyper-parameters in our method, we set $\lambda_1=0.1$ in all experiments. We set $\lambda_2=0.01$ for GTA5 and set $\lambda_2=0.008$ for SYNTHIA.
	All experiments of our method are conducted on a single NVIDIA RTX 3090Ti GPU. 
	
	\subsection{Comparison to State-of-the-Art Methods}
	We compare our method with the state-of-the-art UDA methods on two common UDA benchmarks.
	Depending on whether the distillation technique is used, we divide these methods into two broad categories. Table~\ref{table:gta} and Table~\ref{table:synthia} give the results. From the results, we have the following observations.
	\begin{table}
		\caption{Ablation study on our proposed components.}
		\centering
		\renewcommand{\arraystretch}{1.2}
		\setlength{\tabcolsep}{6pt}
		\begin{tabular}{c|c|c|c|c}
			\toprule
			CMix	&  PiCL  & ProCL	&	HTCM & mIoU (\%)					\\
			\hline\hline
			$\surd$		&		 	    & 			  & 							& 52.1						        \\
			$\surd$		& $\surd$		 	    & 			  & 							& 55.2						        \\
			$\surd$		&		 	    & $\surd$			  & 							& 55.6						        \\	
			$\surd$		&		 	    & 			  & $\surd$							& 56.6						        \\		
			$\surd$		& $\surd$	& $\surd$	& 							& 56.2				    \\
			$\surd$		& $\surd$	& 			  & $\surd$ 			& 57.5				      \\
			$\surd$		& 			  & $\surd$	& $\surd$				& 57.7 				      \\
			$\surd$		& $\surd$	& $\surd$	& $\surd$				& \textbf{58.8}	\\
			\bottomrule	
		\end{tabular}
		\vspace{2mm}	
		
		\label{table:ablation_component}
	\end{table}
	
	% \begin{table}	
		% 	\caption{Ablation study on our proposed components.}
		% 	\centering
		% 	\renewcommand{\arraystretch}{1.2}
		% 	\setlength{\tabcolsep}{8pt}
		% 	\begin{tabular}{cc}
			% 		\toprule
			% 		Methods  			& mIoU (\%)			\\
			% 		\hline
			% 		\hline
			% 		Source Only	& 35.3					\\
			% 		Ours w/o HTCM 		& 56.2					\\
			% 		Ours w/o ProCL		& 57.5					\\
			% 		Ours w/o PiCL		& 57.7					\\
			% 		Ours				& 58.8 					\\
			% 		\bottomrule	
			% 	\end{tabular}
		% 	\vspace{2mm}
		% 	\label{table:ablation_component}
		% \end{table}
	
	First, our method outperforms all non-distillation methods, especially under the SYNTHIA$\rightarrow$Cityscapes setting, where our method outperforms the second-best method by 4.5$\%$ mIoU and 3.4$\%$  mIoU on 16 classes and 13 classes.
	Second, we found that ProDA, the first method using distillation technology, outperforms all previous non-distillation methods. However, our method not only outperforms ProDA, but also a series of subsequent methods based on distillation techniques. For example, our method is better than CRA~\cite{wang2021cross} by $0.2\%$ in GTA5$\rightarrow$Cityscapes and $0.9\%$ in SYNTHIA$\rightarrow$Cityscapes. %Even recent distillation-based methods~\cite{zhang2021multiple,wang2021cross} introduce more sophisticated training techniques
	%our method show competitive performance, i.e., we achieve the similar performance against~\cite{wang2021cross} on GTA5$\rightarrow$Cityscapes and even outperform~\cite{wang2021cross} by $0.9\%$ on SYNTHIA$\rightarrow$Cityscapes. 
	Third, it should be noted that our method not only outperforms the distillation-based methods, but also simplifies the training process. In particular, our method does not require an additional two-step distillation process, nor does it require special training techniques. 
	
	In addition,  we qualitatively compare our model with the source only model ($\mathbb{L}_{s}$) and baseline model ($\mathbb{L}_{s}$+$\mathbb{L}_{st}$) by showing the segmentation results in Figure~\ref{figure:vis}. It can be seen that baseline significantly improves the segmentation results compared to the source only model. But for the thin objects such as "bicycle" (yellow box) and "pole"(green box), almost all pixels are boundary samples, resulting in an incorrect prediction by the baseline model. After using our proposed HTCM and multi-level contrastive loss, many of these mispredicted pixels are corrected back. This demonstrates the effectiveness of our method. 
	
	\subsection{Ablation Study}
	In this section, we conduct experiments to reveal the effectiveness of our proposed method. All experiments are particularly conducted on GTA5$\rightarrow$Cityscapes benchmark.

	\subsubsection{Effect of Components}
	In this section, we validate the individual effects of our proposed HTCM, ProCL, and PiCL. As shown in Table~\ref{table:ablation_component}, our final method achieves an improvement of 6.7\% over the model only using CMix, while removing each one of our components will cause a performance drop comparing to our final method. The results validate that each of our proposed components plays an important role in UDA semantic segmentation.
	
	\begin{table}[htb]
		\caption{Comparison of different mix-up strategies.}
		\centering
		\renewcommand{\arraystretch}{1.2}
		\setlength{\tabcolsep}{8pt}
		\begin{tabular}{l|c}
			
			\toprule
			Methods  				& mIoU (\%)			\\
			\hline
			\hline
			Source Only	& 35.3\\
			DACS					& 52.1				\\
			HTCM					& 52.8	\\
			\hline
			DSP (Hard)		& 53.6 	\\
			DSP (Soft)		& 54.5 	\\	
			DACS+HTCM			& 56.6	 	\\
			\bottomrule	
		\end{tabular}
		\vspace{2mm}
		\label{table:data augmentation}
	\end{table}
	
	\begin{table}[h]
		\def\arraystretch{1.4}
		\setlength{\tabcolsep}{7pt}
		\begin{center}
			\caption{Ablation study of confident threshold $\tau$. A threshold of $\tau \in [0,1)$ corresponds to the confident score of pseudo labeled pixels which is used for GHTB.}
			\vspace{2mm}
			\begin{tabular}{lccccccc} 
				\hline
				
				$\tau \qquad$   & 0.00   & 0.35 & 0.55 & 0.75 & 0.95 &0.97		\\ \toprule 
				mIoU 			& 52.9 	 & 53.8 & 54.6 & 55.2 &  56.6 &  55.8	\\ \bottomrule
				
			\end{tabular}
			\vspace{2mm}
			\label{table:threshold}
		\end{center}
		
	\end{table}
	
	\subsubsection{Comparing Different Mix-up Strategies}
	
	In this section, we compare different mix-up strategies to demonstrate the superiority of HTCM. In particular, we compare DACS~\cite{tranheden2021dacs} and DSP~\cite{gao2021dsp}. Table~\ref{table:data augmentation} gives the results.
	First, compared to DACS, which explicitly constructs target-domain-style context for source boundaries by copying source pixels into a target image, our method HTCM exceeds it by $0.7\%$. This demonstrates the importance of constructing target boundaries with correct labels. In addition, combining HTCM and DACS can further improve performance, proving that our method is orthogonal to DACS.
	Secondly, DSP additionally introduces a mix-up strategy of source-to-source based on DACS. To be fair, we use DACS+HTCM to compare with DSP. In particular, DSP adopts a soft-paste strategy to improve performance. We use hard or soft to distinguish whether it adopts this strategy. It can be seen that the source-to-source strategy used by DSP is inferior to HTCM. This further demonstrates the superiority of our method.

	\subsubsection{Confident Threshold $\tau$}
	In this section, we analyze the impact of different thresholds on performance. In Table~\ref{table:threshold}, we show the results for $\tau=0,0.35,0.55,0.75,0.95,0.97$.
	We find that setting relatively high threshold ($\tau=0.95$) can more effectively boost performance (compared to $\tau=0$). This well shows the importance of constructing the boundaries with high-confidence pixels. However, when we increase the threshold from 0.95 to 0.97, the mIoU decreases from 56.6$\%$ to 55.8$\%$. This is because a too high threshold would filter too many pixels, resulting in insufficient learning.
	\begin{table}[h]
		\def\arraystretch{1.4}
		\setlength{\tabcolsep}{7pt}
		\begin{center}
			\caption{The effect of contrastive learning on different mixed images.}
			\vspace{0.5mm}
			\begin{tabular}{lcc} 
				\hline
				
				Mixed Image    & $x_{hts}$   & $x_{st}$  		\\ \toprule 
				mIoU 			& 56.8 	 & 58.8  	\\ \bottomrule
				
			\end{tabular}
			\vspace{0.5mm}
			\label{table:mixed_image}
		\end{center}
		
	\end{table}
	\subsubsection{Contrastive Learning with Different Images}
	In this section, we compare the performance of $x_{st}$ and $x_{hts}$ for contrastive learning. Table~\ref{table:mixed_image} gives the results. It can be seen that the contrastive learning on $x_{hts}$ is 2 $\%$ lower than the contrastive learning on $x_{st}$. As analyzed in Sec.~\ref{mcl}, almost all pixels in $x_{hts}$ have the correct label, the classification loss at this time provides a good enough supervision signal to learn the discriminative feature representation. As a result, the effect of using contrastive learning to enhance the discriminative of $x_{hts}$ features would become very weak. 
	
	%iffalse
	\subsubsection{Training Cost Discussion}
	Here we particularly compare the training costs of DSP~\cite{gao2021dsp} and ProDA~\cite{zhang2021prototypical} with our method on GTA5$\rightarrow$Cityscapes.   
	Table~\ref{table:training_cost} gives the results. Compared with ProDA and DSP, our method has not only lower training cost but also higher performance. It demonstrates the superiority of our method.
	
	\begin{table}[h]
		\def\arraystretch{1.4}
		\setlength{\tabcolsep}{7pt}
		\begin{center}
			\caption{Training cost for different methods.}
			\vspace{0.5mm}
			\begin{tabular}{lccc} 
				\hline
				
				Method    & GPUs     &Time (hours) &mIoU  		\\ \toprule 
				ProDA 			& 4*Tesla-V100-32GB 	  &108 &57.5  	\\ \bottomrule
				DSP 			& 1*RTX-3090-24GB 	  &96 &55.0  	\\ \bottomrule
				Ours 			& 1*RTX-3090-24GB 	  &63 &58.8  	\\ \bottomrule
				
			\end{tabular}
			\vspace{0.5mm}
			\label{table:training_cost}
		\end{center}
		
	\end{table}
	%fi
	%As mentioned before, We add a confident threshold in our implementation of TMix. This corresponds to only pseudolabeled pixels in which the target output probability exceeds some threshold $\tau$ (other pixels are ignored when mixing). This confident threshold makes sure that we can extract the most likely correct regions in target domain for mixing.
	%In Table~\ref{table:threshold}, we show the results for $\tau=0,0.35,0.55,0.75,0.95$. We find that setting a high confident threshold ($\tau>0$) can effectively boost performance (compared to $\tau=0$). Besides, $\tau=0.95$ achieves state-of-the-art performance on GTA5-to-Cityscapes.
	
	\section{Conclusion}
	In this paper, we design a simple yet effective \emph{non-distillation} framework for UDA semantic segmentation. 
	We propose a novel mix-up strategy to generate high-quality target domain boundary pixels. Specifically, we copy the high-confidence target domain boundary pixels into the source image so that some high-confidence target domain pixels become new boundary pixels. In addition, we propose multi-level contrastive learning to obtain more discriminative feature representations, which contains pixel-to-pixel and prototype-to-prototype contrastive loss.
	We experimentally verified the effectiveness of our proposed methods on GTA5$\rightarrow$Cityscapes and SYNTHIA$\rightarrow$Cityscapes. 
	We believe that our work provides a feasible technical route for designing efficient UDA semantic segmentation methods that bypass the distillation techniques.

	%\section{Acknowledgments}
	
	%Identification of funding sources and other support, and thanks to
	%individuals and groups that assisted in the research and the
	%preparation of the work should be included in an acknowledgment
	%section, which is placed just before the reference section in your
	%document.
	
	%This section has a special environment:
	%\begin{verbatim}
	%  \begin{acks}
		% ...
		%  \end{acks}
	%\end{verbatim}
	%so that the information contained therein can be more easily collected
	%during the article metadata extraction phase, and to ensure
	%consistency in the spelling of the section heading.
	
	%Authors should not prepare this section as a numbered or unnumbered {\verb|\section|}; please use the ``{\verb|acks|}'' environment.

	\begin{acks}
		This work is supported by the National Natural
		Science Foundation of China under Grant 62176246
		and 61836008.
	\end{acks}
	
	\vfill\eject
	\bibliographystyle{ACM-Reference-Format}
	\balance
	\bibliography{sample-base.bbl}
	
	%\appendix

\end{document}